\documentclass[letterpaper, 10 pt, conference]{ieeeconf}  % Comment this line out if you need a4paper

\IEEEoverridecommandlockouts                              % This command is only needed if 
\usepackage{cite}
\usepackage{svg}
\usepackage{amsmath,amssymb,amsfonts}
\usepackage{algorithmic}
\usepackage{graphicx}
\usepackage{graphics}
\usepackage{textcomp}
\usepackage{xcolor}
\usepackage{dsfont}
\usepackage{subfigure}
\usepackage{url}
\usepackage{tabularx,booktabs}

\makeatletter
\let\NAT@parse\undefined
\makeatother
\usepackage{hyperref}

\title{\LARGE \bf
From Semi-supervised to Omni-supervised\\ Room Layout Estimation Using Point Clouds
}

\author{Huan-ang Gao$^{1,2}$, Beiwen Tian$^{1,2}$, Pengfei Li$^{1,2}$, Xiaoxue Chen$^{1,2}$\\ Hao Zhao$^{3,4}$, Guyue Zhou$^{2}$, Yurong Chen$^{4}$ and Hongbin Zha$^{3}$% <-this % stops a space
\thanks{$^{1}$Department of Computer Science and Technology, Tsinghua University, China,
        \{gha20,tbw18,li-pf22,chenxx21\}@mails.tsinghua.edu.cn.}%
\thanks{$^{2}$Institute for AI Industry Research (AIR), Tsinghua University, China,
        zhouguyue@air.tsinghua.edu.cn.}%
\thanks{$^{3}$Peking University, China,  
        zhao-hao@pku.edu.cn, zha@cis.pku.edu.cn.}
\thanks{$^{4}$Intel Labs, China,  
        \{hao.zhao,yurong.chen\}@intel.com.}
}

\begin{document}

\maketitle
\thispagestyle{empty}
\pagestyle{empty}

\renewcommand{\thefootnote}{*}
%%%%%%%%%%%%%%%%%%%%%%%%%%%%%%%%%%%%%%%%%%%%%%%%%%%%%%%%%%%%%%%%%%%%%%%%%%%%%%%%
\begin{abstract}
Room layout estimation is a long-existing robotic vision task that benefits both environment sensing and motion planning. However, layout estimation using point clouds (PCs) still suffers from data scarcity due to annotation difficulty. As such, we address the semi-supervised setting of this task based upon the idea of model exponential moving averaging. 
But adapting this scheme to the state-of-the-art (SOTA) solution for PC-based layout estimation is not straightforward. To this end, we define a quad set matching strategy and several consistency losses based upon metrics tailored for layout quads. Besides, we propose a new online pseudo-label harvesting algorithm that decomposes the distribution of a hybrid distance measure between quads and PC into two components. This technique does not need manual threshold selection and intuitively encourages quads to align with reliable layout points.
Surprisingly, this framework also works for the fully-supervised setting, achieving a new SOTA on the ScanNet benchmark.
Last but not least, we also push the semi-supervised setting to the realistic omni-supervised setting, demonstrating significantly promoted performance on a newly annotated ARKitScenes testing set. Our codes, data and models are made publicly available\footnote{\textcolor{magenta}{\href{https://github.com/AIR-DISCOVER/Omni-PQ}{https://github.com/AIR-DISCOVER/Omni-PQ}}}.
\end{abstract}

%%%%%%%%%%%%%%%%%%%%%%%%%%%%%%%%%%%%%%%%%%%%%%%%%%%%%%%%%%%%%%%%%%%%%%%%%%%%%%%%
\section{Introduction}

Over the past decade, room layout estimation has drawn a lot of attention from the robotics community \cite{ma2016cpa,hsiao2017keyframe,kaess2015simultaneous,yang2016pop,yang2019monocular, zhao2021transferable} since it marks a crucial step towards understanding indoor scenes and might help robot agents make better decisions in challenging environments \cite{chung2022distributed, lasst, perez2021robot, yuan2022situ}. However, the majority of earlier efforts exploit perspective or panoramic RGB images as input \cite{yang2016real, hedau2009recovering, hosseinzadeh2018structure, pintore2020atlantanet, yan20203d, zhao2017physics, zou2021manhattan, schwing2012efficient, liu2015rent3d, zhang20213d, huang2018holistic, hirzer2020smart, zhang2020geolayout, lin2019deeproom, zhao2013scene, fernandez2020corners, solarte2022360, fernandez2018layouts}, whereas the promising paradigm of layout estimation using point clouds (PCs) \cite{chen2022pq} still suffers from the lack of annotated data. It is due to the difficulty of annotating the boundaries of 3D indoor scenes manually, particularly rooms containing non-cuboid shapes and many corners.

We envision an omni-supervised setting \cite{radosavovic2018data} where intelligent robots all over the world can exploit enormous unannotated raw point clouds to continuously improve the collective intelligence (i.e., layout estimation accuracy in this study). To this end, we start from the semi-supervised setting in which we assume a large portion of ScanNet \cite{dai2017scannet} annotations is not available and push it finally to the omni-supervised setting using the recent ARKitScenes \cite{dehghan2021arkitscenes} dataset. 

Actually, semi-supervised room layout estimation has already been studied in a recent work SSLayout360 \cite{tran2021sslayout360}. However, it still relies upon hand-crafted post-processing and only exploits the model exponential moving averaging (EMA) technique to learn representations from many unannotated panoramic images. Note that this paradigm does not apply to the state-of-the-art (SOTA) PC-based layout estimator \cite{chen2022pq}, which directly predicts quads end-to-end.
%To resolve the issue of shortage of annotation, research on semi-supervised learning (SSL) \cite{van2020survey}, which may take use of a wealth of unlabeled data for improving model learning in the context of limited labeled data, has been gaining more and more popularity in recent years. SSLayout360 \cite{tran2021sslayout360} initiatively proposed a semi-supervised indoor layout estimation framework with panoramic images as input, but they didn't investigate the problem in a 3D point cloud version, which is more intriguing and more feasible for application.

\begin{figure}[tbp]
  \centering
  \includegraphics[width=1.00\linewidth]{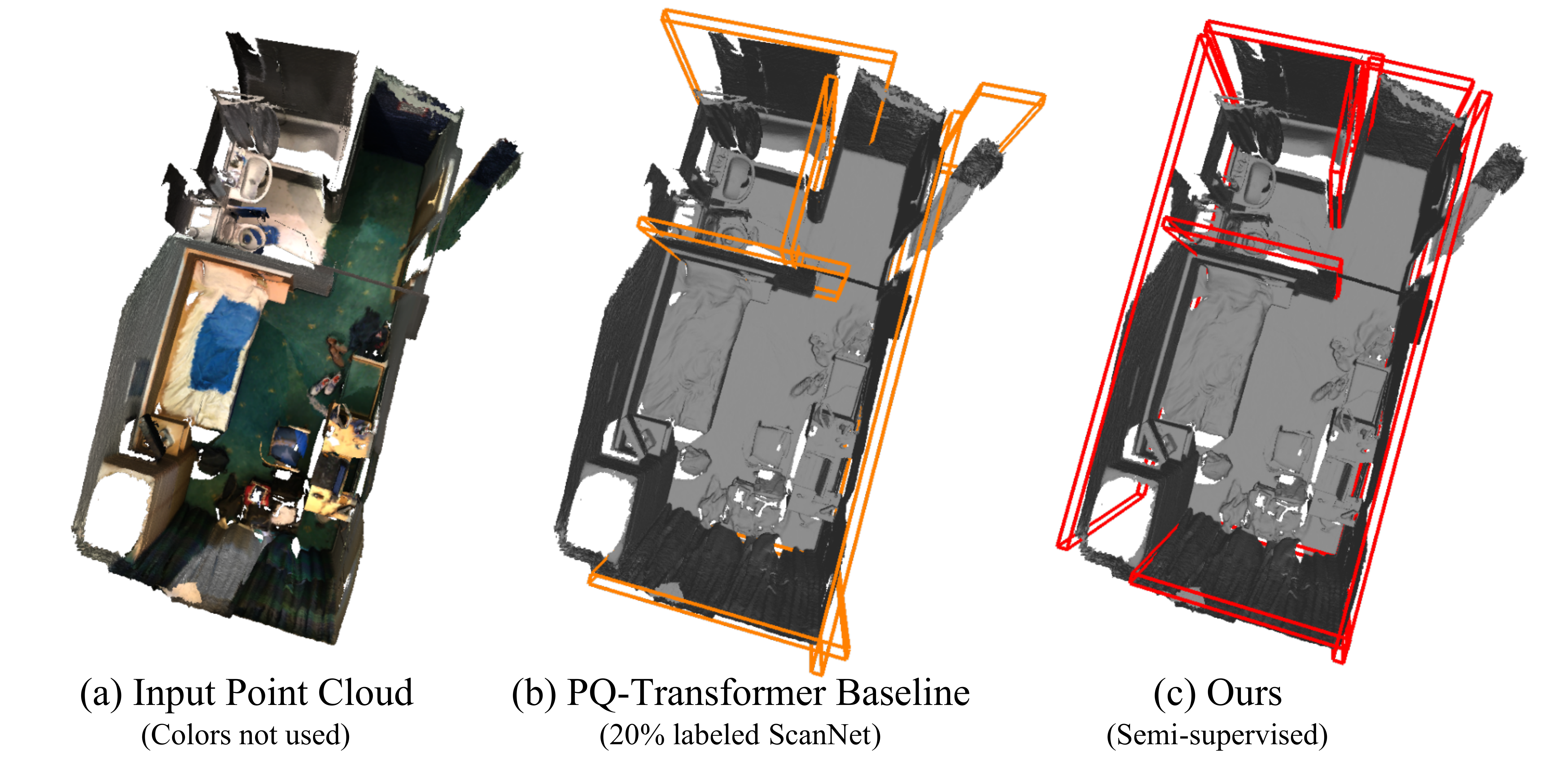}
  \caption{\textit{(a)} The input is a 3D point cloud whose colors are only for visualization. \textit{(b)} We train the former SOTA method PQ-Transformer with only 20$\%$ labeled data of ScanNet training set and use it as the baseline. \textit{(c)} We adopt our method on the whole ScanNet training set with only 20$\%$ annotations, resulting in a more accurate layout prediction.}
  \label{fig:teaser} 
  \vspace{-10pt}
\end{figure}

To this end, we propose the first semi-supervised room layout estimation method using point cloud inputs. Our method builds upon the SOTA counterpart PQ-Transformer \cite{chen2022pq}, which takes the 3D point cloud of a scene as input (see Fig.~\ref{fig:teaser}(a)) and predicts a set of quadrilateral (referred to as \emph{quads}) equations representing layout elements (wall, floor and ceiling). As observed in Fig.~\ref{fig:teaser}(b), it performs poorly on unseen scenes if only $20\%$ annotations are used for training. By contrast, our model is able to predict a more accurate layout by making use of the unlabeled data (see Fig.~\ref{fig:teaser}(c)).

%To this end, we propose our method, which is, to the best of our knowledge, the first approach that trains deep learning models to infer room layouts with point cloud (PC) inputs in a semi-supervised way. Our research follows PQ-Transformer \cite{chen2022pq}, the former state-of-the-art solution for PC-based layout estimation in the fully supervised setting. It takes 3D point cloud of a scene as input (See Fig.~\ref{fig:teaser}(a)) and predicts a set of quadrilateral (referred to as \emph{quads}) equations representing layout elements (wall, floor and ceiling). We further investigate the work in a semi-supervised scenario, namely training with only a tiny ratio of labeled data and a large amount of unlabeled data. As observed in Fig.~\ref{fig:teaser}(b), the layout quad predictions are unable to capture the fine-grained interior corners in the scene because of shortage of labeled training data. However, when our method is adopted, the model is able to predict a more accurate layout making use of the unlabeled data (See Fig.~\ref{fig:teaser}(c)).

Specifically, the success of our method is credited to two techniques. The \textbf{first} is a consistency based training framework inspired by the Mean Teacher \cite{tarvainen2017mean} method. We design a quad matching strategy and three consistency regularization losses that are tailored for the layout estimation problem. We also identify a simple but effective add-on that capitalizes on the confidence of the teacher model. The \textbf{second} is a pseudo label generation module that decomposes the distribution of a new hybrid metric into two components, based upon gamma mixture. It intuitively aligns quad predictions to reliable layout point clouds. Through ablation experiments, both techniques are proven effective, and combining them brings larger improvements. 

%Our approach combines consistency regularization and pseudo-label refinement, two well-known SSL concepts. The consistency regularization requires the model to maintain consistency in face of input data disturbances. We modify the Mean Teacher \cite{tarvainen2017mean} framework to become \textbf{Confident Mean Teacher} (CMT), which fully capitalizes on the confidence of the teacher model on its predictions. A quad set alignment technique is proposed since there is no simple way to measure the consistency of two quad set predictions. And the goal of pseudo-label refinement, or our \textbf{Gamma Mixture Filtering} (GMF) module, is to harvest pseudo-labels of layout quads and refine them into more accurate ones. Ablations demonstrate effectiveness of either of the two proposed methods, and by combining two of them, we see continuing improvement of performance.

Experimental results highlight four notable messages: (1) our solution with different percentages (e.g., $5\%$ to $40\%$) of annotations available consistently and greatly outperforms supervised baselines on the ScanNet dataset. (2) with only 40$\%$ of labeled data we are able to surpass prior fully-supervised SOTA. (3) in the fully-supervised setting, our method can also improve strong baselines by +4.11$\%$. (4) we further extend the method into a more realistic omni-supervised \cite{radosavovic2018data} setting, where we leverage all ScanNet training data and unlabeled ARKitScenes \cite{dehghan2021arkitscenes} training data. On a newly crowd-sourced ARKitScenes testing set, a significant performance gain is achieved, with F1-score going from 10.66$\%$ to 25.85$\%$. Our contributions are as follows:

\begin{itemize}
\item[$\bullet$] We propose the first semi-supervised framework for room layout estimation using point clouds, with tailored designs including a quad set matching strategy and three confidence-guided consistency losses.
\item[$\bullet$] We propose a threshold-free pseudo-label harvesting technique based upon a newly-proposed hybrid distance metric and gamma mixture decomposition.
\item[$\bullet$] We achieve significant results in semi-supervised, fully-supervised and omni-supervised settings. We contribute a new crowd-sourced testing set and release our codes.
\end{itemize}

\section{Related Works}
Recently, semi-supervised and weakly-supervised learning are hot topics in the robotics community, with many methods proposed for various tasks including point cloud semantic parsing \cite{liu2022weaklabel3d, deng2022superpoint, tian2022vibus, yi2021complete} and representation learning \cite{huang2021spatio}, 3D object detection \cite{ding2022jst}\cite{zhang2022atf}, articulation understanding \cite{lv2022sagci}, single-view reconstruction \cite{li2021monocular} and intrinsic decomposition \cite{wang2021semi}. This line of research envisions an exciting future scheme that robots all over the world exploit unlimited unlabeled data to continuously improve the collective intelligence. \cite{zhao2022codedvtr,li2022hdmapnet,zhang2022mutr3d,liu2022hoi4d} Our study is the first semi-supervised framework for room layout estimation from point clouds, which contributes to this robotic vision trend.

From the perspective of methodology, we briefly review two kinds of semi-supervised learning (SSL) paradigms. 

\textbf{The consistency based SSL methods} rely on the assumption that near samples from the low-dimensional input data manifold result in near outputs in the feature space \cite{ghosh2020data}\cite{van2020survey}. Thus, they enforce the model to stay in agreement with itself despite perturbations. Under this scope, multiple perturbation strategies are explored.
The $\Pi$ model \cite{rasmus2015semi}\cite{laine2017temporal} penalizes the difference of hidden features of the same input with different data transformations and dropout.
Temporal Ensembling training \cite{laine2017temporal} regularizes consistency on current and former predictions.
The Mean Teacher method \cite{tarvainen2017mean} uses exponential moving average of student network parameters. 

\textbf{The pseudo-label based SSL methods}, on the other hand, are more general as they don't require domain-specific data transformations.
By equipping them with a few necessary designs, they can be as proficient as consistency based ones. For example, in 3D object detection task, \cite{yin2022semi} proposes two post-processing modules to improve the recall rate and the precision rate of the pseudo labels. In image classification task, \cite{sohn2020fixmatch} sets a constant confidence threshold $\tau$ for determining whether to discard a pseudo-label, and \cite{zhang2021flexmatch} upgrades that constant to a set of per-class learnable variables. In 2D object detection task, Noisy Pseudo-box Learning strategy is proposed by \cite{li2022pseco}, which only considers $N$ proposals of top-quality as pseudo labels and the rest ones as false positives.

%By means of combining the consistency based methods and the pseudo-label based methods together, state-of-the-art results have been achieved by many works. [TODO: ref more works here] As far as we know, there are no prior works focused on layout estimation task in a semi-supervised setting using PCs only. Therefore, here we present our PC-based semi-supervised layout estimation method, with a \textbf{Confident Mean Teacher} module regularizing consistency and a \textbf{Gamma Mixture Filtering} module refining pseudo-labels.

\begin{figure*}[htbp]
  \centering
  \includegraphics[width=0.83\linewidth]{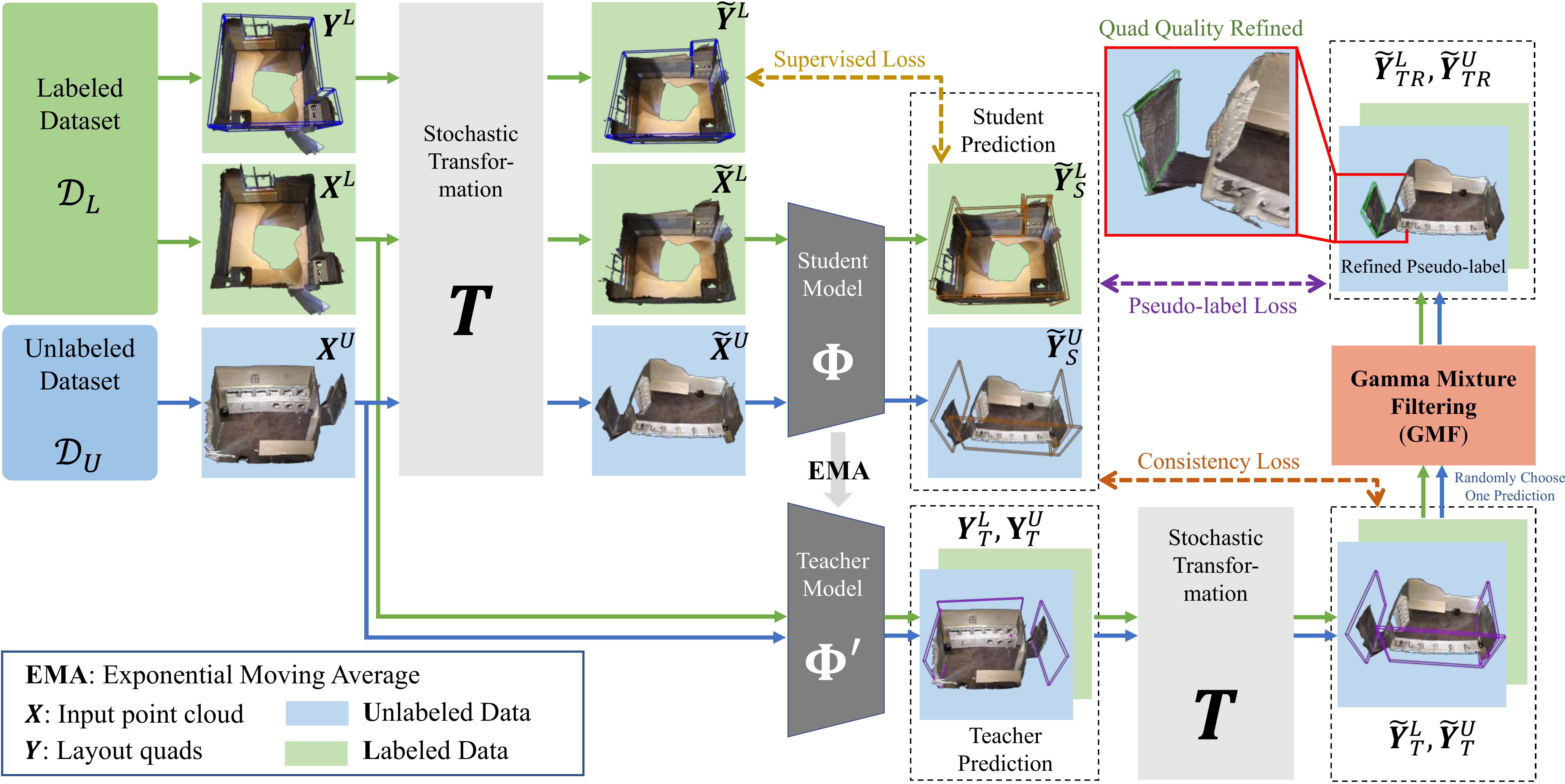}
  \caption{\textbf{Method Overview}. In each training iteration, we sample $(\mathbf{X}^L, \mathbf{Y}^L)$ from labeled dataset and $\mathbf{X}^U$ from unlabeled dataset to form a batch.
  The input batch is first stochastically transformed then fed into the student model to produce predictions $\tilde{\mathbf{Y}}^L_S$ and $\tilde{\mathbf{Y}}^U_S$.
  Meanwhile, the input batch is also fed into the teacher model then transformed to produce predictions $\tilde{\mathbf{Y}}^L_T$ and $\tilde{\mathbf{Y}}^U_T$.
  In the two adopted transformations, FPS sampling uses different seeds whereas rotation, flipping and scaling are identical.
  We impose three losses in total:
  (1) a supervised loss between the transformed label and predictions of student model. 
  (2) a consistency loss that minimizes the difference between student predictions and teacher predictions.
  (3) a pseudo-label loss that encourages quads to align with reliable layout points.
  The student parameters are updated by gradient descent according to the sum of three losses, whereas the teacher parameters are updated by exponential moving average (EMA) of student parameters.}
  \label{fig:main} 
  \vspace{-10pt}
\end{figure*}

\section{Method}
We aim to develop a learning framework that allows robot agents to leverage enormous unlabeled data to infer room layouts $\mathbf{Y}$ from indoor scene point clouds $\mathbf{X}$.
Following \cite{chen2022pq}, we denote a layout wall (represented by a quad) $\mathbf{y} = \{\mathbf{c}, \mathbf{n}, \mathbf{s}, p\} \in \mathbf{Y}$ by its center coordinate $\mathbf{c}$, unit normal vector $\mathbf{n}$, size $\mathbf{s} = (w, h)$, and predicted quadness score $p$. 
The quadness scores of ground truths are fixed to $1.0$.

% For the $i$-th scene, we denote point coordinates by $\mathbf{X}_i$ and the ground-truth annotation by $\mathbf{Y}_i$.

To start with, we formally describe three training settings.
Suppose we have a 3D point cloud (PC) dataset $\mathcal D_L$ with layout annotations in conjunction with a much larger unlabeled PC dataset $\mathcal D_U$.
In the \textbf{fully-supervised setting}, $\mathcal{D}_L$ is the whole training set of ScanNet with quad annotations whereas $\mathcal{D}_U$ is a null set.
In the \textbf{semi-supervised setting}, $\mathcal{D}_L$ is part of the ScanNet training set along with quad annotations whereas $\mathcal{D}_U$ is the complementary set whose annotations are assumed unknown. In the \textbf{omni-supervised setting} \cite{radosavovic2018data} which is a real-world generalization of the semi-supervised setting,
$\mathcal{D}_L$ is the whole training set of ScanNet with annotations whereas $\mathcal{D}_U$ is the ARKitScenes training set without annotations.

We introduce our method in the three settings using unified notations $\mathcal{D}_L$ and $\mathcal{D}_U$.
% We introduce our method under the semi-supervised setting because the other two are special cases of it.
% For the $i$-th scene with $N_i$ points and $K_i$ layout walls (or quads), we use $\bold x_i \in \mathbb R^{N_i \times 3}$ to represent coordinates of $N_i$ point coordinates and $\bold y_i \in \mathbb R^{K_i \times (3+3+2+1)}$ to represent corresponding $K_i$ scene layout walls (or quads).
As depicted in Fig. \ref{fig:main}, we adapt the Mean Teacher \cite{tarvainen2017mean} training framework (see Sec. \ref{CMT}) to end-to-end room layout estimation with a tailored quad matching strategy and three consistency losses. We also integrate a novel pseudo-label refinement module (Sec. \ref{GMF}) for quads, which is based upon gamma mixture decomposition.
%In , we describe the Confident Mean Teacher (CMT) framework and its improvements compared with the typical Mean Teacher.
% and also define the loss terms $\mathcal{L}_{\text{sup}}$ and $\mathcal{L}_{\text{CMT}}$.
%In , we introduce the Gamma Mixture Filtering (GMF) module which aims to refine the quad predictions, so that the re-estimated quads serve as pseudo-labels with higher accuracy.
In Sec. \ref{LOSS}, we describe the loss terms to optimize.

\subsection{Quad Mean Teacher (QMT)}

\label{CMT}

% \textbf{Review of Mean Teacher.}
Mean Teacher \cite{tarvainen2017mean} is a successful framework for semi-supervised learning with a student model and a teacher model of the same architecture.
The general idea is to feed two models with the same input samples transformed differently and enforce the predictions of the two models to be consistent.
The student model is updated by gradient descent while the teacher model is updated by exponential moving average (EMA) of the weights of the student model.

% In each training iteration of our method, we sample $\mathbf{x}^L$ from $\mathcal D_L$ and $\mathbf{x}^U$ from $\mathcal D_U$ to form a batch $\mathbf{x} = \{\mathbf{x}^L, \mathbf{x}^U\}$.
Inspired by the idea of Mean Teacher, we first sample $\mathbf{X}^U$ from $\mathcal D_U$ and $(\mathbf{X}^L, \mathbf{Y}^L)$ from $\mathcal D_L$ to form a batch $\mathbf{X} = \{\mathbf{X}^L, \mathbf{X}^U\}$.
$\mathbf{X}$ is transformed with stochastic transformation $T$ before feeding into the student model to yield $\tilde{\mathbf{Y}}_S = \{\tilde{\mathbf{Y}}_S^L, \tilde{\mathbf{Y}}_S^U \}$.
$\mathbf{Y}^L$ is transformed into $\tilde{\mathbf{Y}}^L$ with the same transformation.
% $\mathbf{X}$ and $\mathbf{Y}_L$ are transformed with random transformation $\mathbf{\mathcal{T}}$ and then fed into student model to give $\tilde{\mathbf{Y}}_S = \{\tilde{\mathbf{Y}}_S^L, \tilde{\mathbf{Y}}_S^U \}$ and $\tilde{\mathbf{Y}}_L$, 
Meanwhile, $\mathbf{X}$ is also fed into the teacher model and then applied the same transformation $T$ to yield $\tilde{\mathbf{Y}}_T = \{\tilde{\mathbf{Y}}_T^L, \tilde{\mathbf{Y}}_T^U \}$.
Following the same loss design in \cite{chen2022pq}, we impose a supervised loss $\mathcal{L}_{\text{sup}}$ between $\tilde{\mathbf{Y}}_S^L$ and $\tilde{\mathbf{Y}}^L$.

% In the meantime, we aim to enforce the consistency between the quad set predictions of the student and the teacher model.

The success of Mean Teacher based methods relies on domain-specific data transformation and carefully designed consistency losses between two sets of predictions, without which the method could suffer from degeneration.
Based upon this observation, we design the transformation domain and consistency losses for room layout estimation as follows. 
% Among all of the consistency regularization methods, Mean Teacher \cite{tarvainen2017mean} showed its remarkable success.
% The general idea is to keep a teacher model during training, whose parameters are updated using the exponential moving average (EMA) of parameters of its student counterpart.
% Under different distortions of input data, the student model is enforced to learn to remain consistent with teacher via gradient descent.
% Therefore, in the following we discuss our data transformation steps and loss terms in detail.

\textbf{Data transformation}\quad
We adopt four kinds of transformations: Farthest Point Sampling (FPS) \cite{qi2017pointnet++}, flipping along horizontal axes, rotating along vertical axes and coordinates scaling. 
FPS \cite{qi2017pointnet++} downsamples the point cloud by repeatedly choosing the point farthest from the chosen ones, discarding only redundant points.
Also, flipping, rotating and scaling in constrained ways mimic the natural viewpoint changes of humans. 
Among them, layout annotations are invariant to FPS \cite{qi2017pointnet++} as subsampling does not change the layout geometries and equivariant to the other three transformations with which the geometries should be transformed accordingly. 
Hence, when applying the same transformation, for invariant transformation (i.e., FPS \cite{qi2017pointnet++}) we use different seeds and for the other three we apply the same transformation before the student model and after the teacher model.

\begin{figure}[t]
  \centering
  \includegraphics[width=0.83\linewidth]{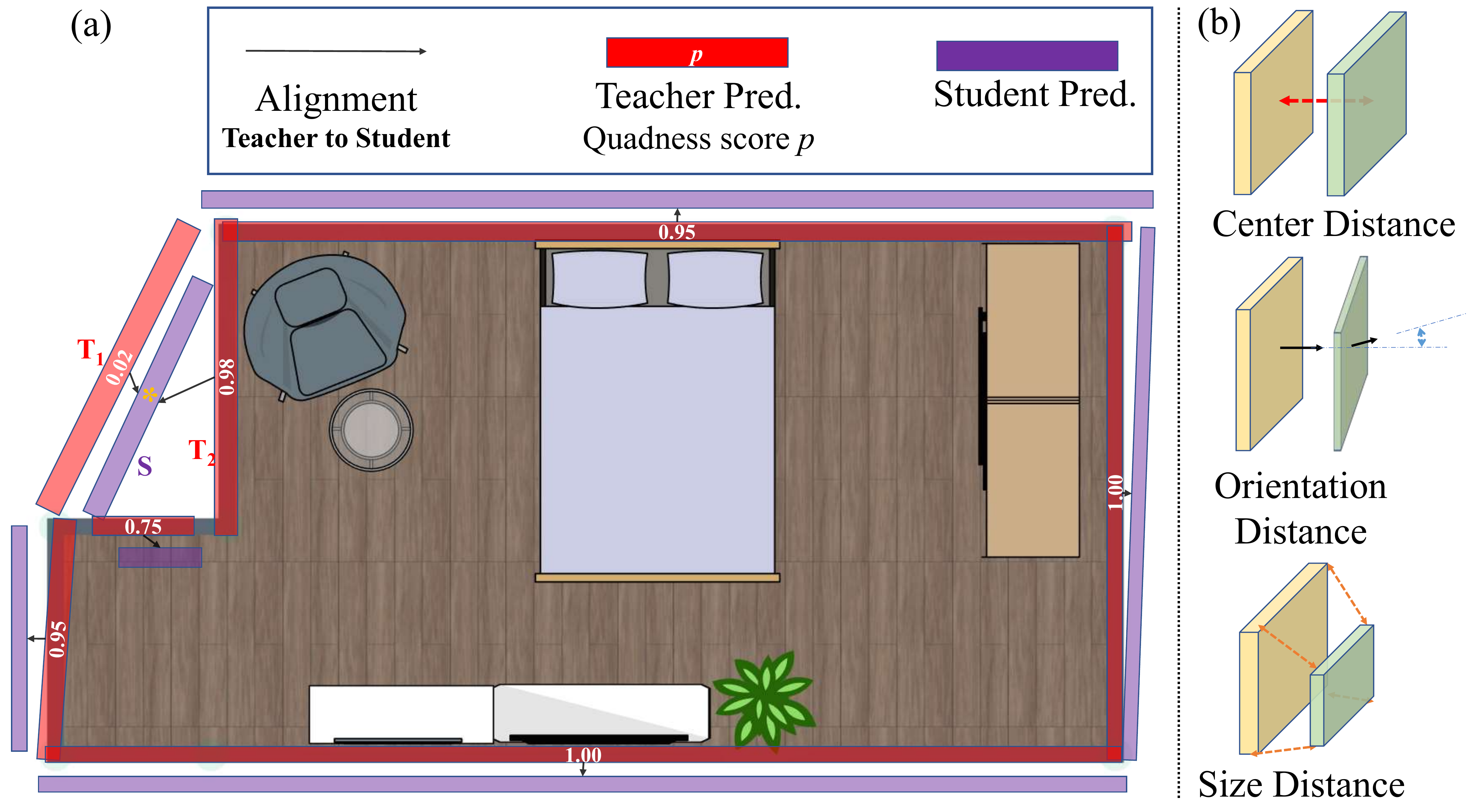}
  \caption{\textbf{Illustration on Teacher Student Alignment}. \textit{(a)} For every teacher-predicted quad, we find the nearest student-predicted quad. Although teacher predictions are noisy, the quadness scores demonstrate how accurate the predictions are. \textit{(b)} These three figure illustrate the three components of the defined distance between two quads.}
  \label{fig:cmt}
  \vspace{-10pt}
\end{figure}

\textbf{Quad Set Matching}\quad
% As aforementioned, there are no direct ways to measure the difference between two quad sets.
To encourage consistency between the predicted quad sets of two models, the difference between two quads should be defined first.
Given two quad predictions, $\tilde{\mathbf{y}}_1 = \{ \tilde{\mathbf{c}}_1, \tilde{\mathbf{n}}_1, \tilde{\mathbf{s}}_1, p_1 \}$ and $\tilde{\mathbf{y}}_2 = \{ \tilde{\mathbf{c}}_2, \tilde{\mathbf{n}}_2, \tilde{\mathbf{s}}_2, p_2 \}$, the differences of three geometrical characteristics (quad center location $\tilde{\mathbf{c}}$, quad normal $\tilde{\mathbf{n}}$, quad size $\tilde{\mathbf{s}}$) should all be considered.
Thus, as illustrated in Fig.~\ref{fig:cmt}(b), we define the distance between two quads as: ($|\!|\cdot|\!|_k$ denotes $k$-norm)
\begin{align}
% d_{\text C}(\tilde{\mathbf{y}}_1, \tilde{\mathbf{y}}_2) &= {|| \tilde{\mathbf{c}}_1 -  \tilde{\mathbf{c}}_2|| } \\
% d_{\text N}(\tilde{\mathbf{y}}_1, \tilde{\mathbf{y}}_2) &= {|| 1 - \tilde{\mathbf{n}}_1 \cdot  \tilde{\mathbf{n}}_2||} \\
% d_{\text S}(\tilde{\mathbf{y}}_1, \tilde{\mathbf{y}}_2) &= {|| \tilde{\mathbf{s}}_1 -  \tilde{\mathbf{s}}_2|| } \\
d(\tilde{\mathbf{y}}_1, \tilde{\mathbf{y}}_2) =
{|\!| \tilde{\mathbf{c}}_1 -  \tilde{\mathbf{c}}_2|\!|_2 } +
{| 1 - \tilde{\mathbf{n}}_1 \cdot  \tilde{\mathbf{n}}_2|} + 
{|\!| \tilde{\mathbf{s}}_1 -  \tilde{\mathbf{s}}_2|\!|_2^2 }
\end{align}

Based on the distance metric between quads, we calculate the difference of two predicted quad sets by first finding the correspondences between the two quad sets and then summing up the distances between corresponding quads.
% The correspondences are establish by a unidirectional mapping from the teacher predictions to student predictions.
To establish the correspondences, we find the nearest student-predicted quad $\tilde{\mathbf{y}}_S = \{ \tilde{\mathbf{c}}_S, \tilde{\mathbf{n}}_S, \tilde{\mathbf{s}}_S, p_S \}$ for each teacher-predicted quad $\tilde{\mathbf{y}}_T = \{ \tilde{\mathbf{c}}_T, \tilde{\mathbf{n}}_T, \tilde{\mathbf{s}}_T, p_T \}$:
\begin{align}
\mathcal{P}_{\tilde{\mathbf{Y}}_S}(\tilde{\mathbf{y}}_T) = \operatorname*{argmin}_{\tilde{\mathbf{y}}_S \in \tilde{\mathbf{Y}}_S} |\!| \tilde{\mathbf{c}}_S - \tilde{\mathbf{c}}_T |\!|_2
\end{align}
We use $\mathcal{P}(\cdot)$ to represent this injective mapping from the teacher model prediction to the student model prediction.

\textbf{Consistency Loss Design}\quad
Although the quad geometries (i.e., $\tilde{\mathbf{c}}, \tilde{\mathbf{n}}, \tilde{\mathbf{s}}$) predicted by teacher are not adequately precise, the predicted quadness score $p$ could measure the correctness of the predictions. 
Considering that the teacher-predicted quads are generally more reliable than the student-predicted quads, we use teacher-predicted quadness scores $p_T$ as the confidence and define the consistency loss $\mathcal{L}_{\text{QMT}}$ as:
\begin{align}
    \mathcal{L}_{\text{QMT}} = \dfrac{1}{|\tilde{\mathbf{Y}}_T|} \sum_{\tilde{\mathbf{y}}_T \in \mathbf{Y}_T}& d(\mathcal{P}_{\tilde{\mathbf{Y}}_S}(\tilde{\mathbf{y}}_T), \tilde{\mathbf{y}}_T)\cdot p_T 
    % \mathcal{L}_{\text{CMT}} =& D(\tilde{{\mathbf{Y}}}_S,\tilde{{\mathbf{Y}}}_T)
\end{align}

\textbf{Remark}\quad A similar idea to evaluate the closeness of two sets is the Chamfer Distance, which establishes two injective mappings from each of the two sets to its counterpart.
On the contrary, our method only establishes a one-way mapping from the teacher model predictions to the student model predictions since the latter is less reliable than the former.
As depicted in Fig.~\ref{fig:cmt}(a), finding the nearest teacher prediction around $S$ and penalizing the quad distance in between would wrongly push $S$ to the unreliable prediction $T_1$.
By contrast, as $S$ is the nearest student prediction around unreliable $T_1$ and reliable $T_2$, optimizing the weighted quad distance sum would push $S$ to the reliable prediction $T_2$.

% noisy prediction $T_1$ and accurate prediction $T_2$ both corresponds to student prediction $S$ for further optimization.
% If we take the opposite approach, find the nearest teacher prediction to student prediction $S$, then we'll teach $S$ to the wrong target $T_1$.

% Here for simplicity we use $\mathcal{P}(\cdot)$ for representing the operator that maps the teacher prediction to its student counterpart, i.e. $\tilde{\mathbf{y}}_T^{(i)} \xrightarrow{\mathcal{P}} \tilde{\mathbf{y}}_S^{(j)}$. 
% \begin{align}
% \mathcal{P}(\tilde{\mathbf{y}}_T^{(i)}) &= \operatorname*{argmin}\limits_{
% % argmin
% \tilde{\mathbf{y}}_S^{(j)} \in \tilde{\mathbf{y}}_S
% } || \tilde{\mathbf{c}}_S^{(j)} - \tilde{\mathbf{c}}_T^{(i)} ||
% \end{align}

\subsection{Gamma Mixture Filtering (GMF)}

\label{GMF}
% \vspace{-10pt}

% \textbf{Insight.}
% The question posed here is that given a predicted quad $\tilde{\mathbf{y}}_T^{(i)}$, how can we refine it to a more accurate one $\tilde{\mathbf{y}}_{TR}^{(i)}$.
In this stage, we introduce the Gamma Mixture Filtering module which makes further use of the unlabeled data and re-estimates a more accurate quad prediction $\tilde{\mathbf{y}}_{TR}$ from the noisy prediction $\tilde{\mathbf{y}}_T$.
A naive approach to do this is to select points whose perpendicular distance to the quad is below a manually chosen distance threshold $\epsilon_D$ and use these points to estimate a more accurate quad.
However, it is inevitable to manually tune the hyper-parameter $\epsilon_D$, which is time-consuming and ineffective as a fixed threshold is usually not applicable to all scenes.
Besides, using perpendicular distance solely as the metric may erroneously select points in the room corners which belong to other quads.

% But several problems remain to be solved.
% First, the distance metric between a point and a quad is unclear.
% Then, how to select points without using the manually selected thresholds.
% And finally, how to estimate a new quad using selected points.

To address these issues, we introduce 1) hybrid distance between point and quad as an improved metric and 2) the gamma mixture decomposition filtering strategy to automatically select the threshold for filtering.

% $\tilde{\mathbf{y}}_T^{(i)} = \{ \tilde{\mathbf{c}}_T^{(i)}, \tilde{\mathbf{n}}_T^{(i)}, \tilde{\mathbf{s}}_T^{(i)}, p_T^{(i)} \}$

\textbf{Hybrid Point-Quad Metric}\quad
We propose a hybrid metric to measure the distance between a point and a quad.
Instead of using the perpendicular distance alone, we also leverage normals and quad sizes.
Consider a point $\mathbf{p}$ with coordinate $\mathbf{c}_p$ and normal $\mathbf{n}_p$ estimated with adjacent points in the PC, and a quad $\tilde{\mathbf{y}}_T$ whose plane equation is $\tilde{\mathbf{n}}_T \cdot (\mathbf{c} - \tilde{\mathbf{c}}_T) = 0$, $\mathbf{c} \in \mathbb{R}^3$.
Then the perpendicular distance can be written as:
\begin{align}
    \mathcal{M}_\text{p}(\mathbf{p}, \tilde{\mathbf{y}}_T) = {|(\mathbf{c}_p - \tilde{\mathbf{c}}_T) \cdot \tilde{\mathbf{n}}_T|}
\end{align}

Note that $\tilde{\mathbf{n}}_T$ is of unit length.
In some corner cases where points are close but differ greatly in normals (e.g. in wall corners), using this measure solely would erroneously include points on other quads. Therefore, we also define a cosine similarity metric for the normals:
\begin{align}
    \mathcal{M}_\text{o}(\mathbf{p}, \tilde{\mathbf{y}}_T) = |1 - {\mathbf{n}_p \cdot \tilde{\mathbf{n}}_T}|
\end{align}

Furthermore, as the size of quads is not considered in the proposed two measures, we consider the extent to which the projections of points lay outside the quad.
Since the vertical edges of predicted quads are parallel to $\hat{\mathbf{z}} = (0, 0, 1)^T$, the horizontal edges should be parallel to $\hat{\mathbf{x}} = \dfrac {\tilde{\mathbf{n}}_T \times \hat{\mathbf{z}}}{|\!|\tilde{\mathbf{n}}_T \times \hat{\mathbf{z}}|\!|_2}$ ($\times$ denotes cross product).
The horizontal and vertical distances between the quad center and the projection of $\mathbf{p}$ on the quad are then given by $w_p = |(\mathbf{c}_p - \tilde{\mathbf{c}}_T) \cdot \hat{\mathbf{x}}|$ and $h_p = |(\mathbf{c}_p - \tilde{\mathbf{c}}_T) \cdot \hat{\mathbf{z}}|$, respectively.
Thus, we define the out-of-quad metric as:
\begin{align}
    \mathcal{M}_\text{s}(\mathbf{p}, \tilde{\mathbf{y}}_T) = |\!|\text{ReLU}((w_p, h_p)^T - \tilde{\mathbf{s}}_T)|\!|_1
\end{align}

Finally, the hybrid point-quad distance is defined as:
\begin{align}
    % \mathcal{M}(\mathbf{p}, \tilde{\mathbf{y}}_T^{(i)}) = \mathcal{M}_\text{v}(\mathbf{p}, \tilde{\mathbf{y}}_T^{(i)}) + \mathcal{M}_\text{n}(\mathbf{p}, \tilde{\mathbf{y}}_T^{(i)}) + \mathcal{M}_\text{s}(\mathbf{p}, \tilde{\mathbf{y}}_T^{(i)})
    \mathcal{M} = \mathcal{M}_p + \mathcal{M}_o + \mathcal{M}_s
\end{align}

\begin{figure}[t]
  \centering
  \includegraphics[width=1.00\linewidth]{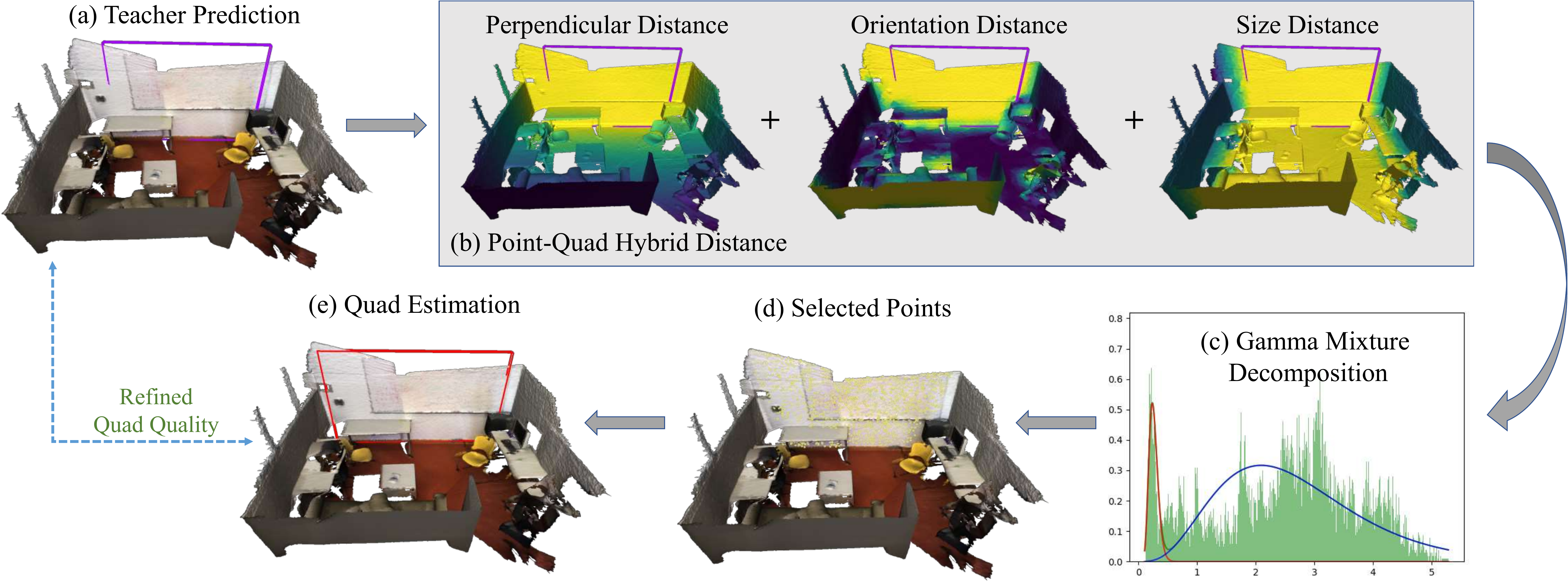}
  \caption{\textbf{Illustration on Gamma Mixture Filtering}. We calculate the proposed hybrid metrics between points and quads in (b), where warmer colors indicate shorter distances. Then we decompose the distribution of metrics into two components, corresponding to points that belong to the quad and those don’t, respectively. We filter out redundant points using the mixture distribution model (depicted in (c)), and re-estimate quads with higher accuracy for the student model to learn.}
  \label{fig:mixture} 
  \vspace{-10pt}
\end{figure}

% The visualization of the three proposed metric is in Fig.~\ref{fig:mixture}(b), illustrating the metrics between the highlighted quad and points.
In Fig.~\ref{fig:mixture}(b) we illustrate the three proposed metrics between the highlighted quad and points.

\textbf{Mixture Decomposition Filtering}\quad
In this stage we use the hybrid metric $x=\mathcal{M}(\cdot, \cdot)$ to select points from the PC for each quad.
We first collect the metrics between the quad and all points, and then use the metrics to fit a probabilistic mixture model.
The possibility density function (PDF) of the probability model is defined as
\begin{align}
    P(x | \theta_0, \theta_1) = w_0 P(x | \theta_0) + w_1 P(x | \theta_1)
\end{align}
where $P(x|\theta_0)$ and $P(x | \theta_1)$ are PDFs of individual components and $w_0, w_1$ denote weights of them, with $w_0 + w_1 = 1$.

The two individual components correspond to points that belong to the quad and those don't, respectively.
We empirically choose gamma distribution for the two components:
\begin{align}
    P(x|\theta_i) = \dfrac{\Gamma(a+b)}{\Gamma(a)\Gamma(b)}x^{a-1}(1-x)^{b-1}, \theta_i = \{a, b\}
\end{align}

To fit this mixture distribution, we follow \cite{zhao2020pointly} to decide the parameters $\theta_0, \theta_1, w_0$ and $w_1$.
By using the expectation maximization (EM) algorithm, we take the parameters when $\sum_{p \in P} \log P(x_p | \theta_0, \theta_1)$ is maximized.
The fitting result is illustrated in Fig.~\ref{fig:mixture}(c), where the blue curve represents $P(x | \theta_0)$ and the red curve represents $P(x | \theta_1)$.

% These two components represent points that belong to the quad and those don't, respectively.
Finally, with this mixture model, we examine the probabilities that an unlabeled point belongs and not belongs to the quad.
When the former is larger than the latter, we keep this point during filtering. 
In other words, for each quad $\mathbf{y}_T$ we keep points $\mathbf{p}_i$ that satisfy $\ w_0 P(x_i | \theta_0) \le w_1 P(x_i | \theta_1)$ where $x_i = \mathcal{M}(\mathbf{p}_i, \mathbf{y}_T)$, as shown in Fig.~\ref{fig:mixture}(d). It is unnecessary to manually tune a threshold, as the intersection point of the two component PDFs works as a per-quad threshold obtained by statistics of the unlabeled points around that quad.
% We define the metric of points by a random variable $X$ subjecting to a mixture distribution consisting of two individual components $\Gamma_{\text{1}}$ and $\Gamma_{\text{0}}$, which represent points that belong to the quad and those don't respectively.

% And that naturally give PCs belonging to quad $q$ can be judged by:
% \begin{align}
%      x_i = \mathcal{M}(p_i, q) \ \text{and} \ w_0 P_0(x_i | \theta_0) \le w_1 P_1(x_i | \theta_1)\label{eq:criterion}
% \end{align}
% Here $P$ denotes the whole PC set.

% Finally we select points that satisfy $\ w_0 P_0(x_i | \theta_0) \le w_1 P_1(x_i | \theta_1)$ where $x_i = \mathcal{M}(p_i, q)$, as shown in Fig.~\ref{fig:mixture}(d).

\textbf{Quad estimation}\quad
With the set of selected points $P'$, we reconstruct a more accurate quad $\tilde{\mathbf{y}}_{TR}$ for each predicted quad $\tilde{\mathbf{y}}_{T}$.
We refine the quad center and quad normal to $
\mathbf{c}' = \frac{1}{|P'|}\sum_{p \in P'} \mathbf{c}_p$ and $\mathbf{n}' = {\sum_{p \in P'} \mathbf{n}_p} \ /\  {|\!|\sum_{p \in P'} \mathbf{n}_p|\!|_2}$.
To estimate the quad size, we randomly take $K_s$ samples $\{\tau_i\}_{i=1}^{K_s}$ from $[0, 1]$.
Under the assumption that the point collection $P'$ is uniformly sampled from the refined quad, we refine the quad size to $\mathbf{s}' = \frac{1}{K_s} \sum_{i=1}^{K_s} \frac{1}{\tau_i} \cdot \rm{quantile}(\tau_i)$.
Here ${\text {quantile}}(\tau_i)$ is defined as $\tau_i$-th quantiles of $\{\mathbf{s}_p | p \in P'\}$ computed on $\hat{\mathbf{x}}$ axis and $\hat{\mathbf{z}}$ axis, respectively.

In each scene of each training step, due to tractability concerns, we choose one of all teacher predicted quads to refine, as illustrated in Fig.~\ref{fig:main}.
Based on the refined quad $\tilde{\mathbf{y}}_{TR} = \{ \mathbf{c}', \mathbf{n}', \mathbf{s}', 1.0 \}$, we propose the pseudo-label loss:
\begin{align}
    \mathcal{L}_\text{GMF} = d(\mathcal{P}(\tilde{\mathbf{y}}_{T}), \tilde{\mathbf{y}}_{TR})
\end{align}

% \subsection{Loss functions}
% \label{loss}
% Training loss is a combination of original supervised loss, consistency loss and pseudo-label loss. That gives:
% \begin{align}
%     \mathcal{L} = \mathcal{L}_\text{sup} + \lambda_\text{CMT} \mathcal{L}_\text{CMT} + \lambda_\text{GMF} \mathcal{L}_\text{GMF}
% \end{align}

% And after student updates its parameter at step $t$, we update teacher parameter by:
% \begin{align}
%     \Phi'_{t} = \alpha \Phi'_{t-1} + (1 - \alpha) \Phi_{t}
% \end{align}

\subsection{Loss}
\label{LOSS}
The loss term we aim to optimize during training is:
\begin{align}
\mathcal{L} = \mathcal{L}_{\text{sup}} + \lambda_{\text{QMT}}\mathcal{L}_{\text{QMT}} + \lambda_{\text{GMF}}\mathcal{L}_{\text{GMF}}
\end{align}
where $\lambda_{\text{QMT}}$ and $\lambda_{\text{GMF}}$ are loss weights.
\section{Experiment}

%In  this  section,  we  first provide our joint detection results of 3D objects and layouts on dataset, and compare them with previous state-of the art methods. Then we show ablation study results to verify the efficiency of physical constraint loss and different proposal methods. Finally we show qualitative results of our approach. 

% \subsection{Comparisons with Fully-supervised methods}
\subsection{Datasets and Implementation Details}
\textbf{Datasets}\quad In the semi-supervised setting, our methods are evaluated on the ScanNet dataset.
ScanNet \cite{dai2017scannet} is a large-scale RGB-D video dataset with 3D reconstructions of indoor scenes, including 1513 scans reconstructed from around 2.5 million views.
On top of the ScanNet, SceneCAD \cite{avetisyan2020scenecad} provides scene layout annotations containing 8.4K polygons.
In our experiments, we use the 3D reconstructions from ScanNet \cite{dai2017scannet} as the input point clouds and use the scene layouts from SceneCAD \cite{avetisyan2020scenecad} as the ground truth labels.

% We introduce a new richly-annotated real-world scene layout dataset consisting of 1151 CAD shells and wireframes on top of the ScanNet RGB-D dataset, allowing large-scale data-driven training for layout estimation.
% Based on ScanNet V2, SceneCAD \cite{avetisyan2020scenecad} proposed a    

% First we choose ScanNet v2 dataset \cite{dai2017scannet} which reconstructs real-world rooms to 3D RGB-D meshes, contributing $\sim$1.2K training scans and $\sim$0.3K validation scans in total. We sample input point clouds using farthest point sampling (FPS) algorithm from vertices of the scanned meshes. For layout quad annotation, we follow SceneCAD \cite{avetisyan2020scenecad}, which introduce $\sim$13.8K corners, $\sim$20.5K edges and $\sim$8.4K polygons to the typical ScanNet training and validation data split.

Furthermore, we extend our methods to the omni-supervised setting and employ ARKitScenes dataset \cite{dehghan2021arkitscenes}.
ARKitScenes is another large-scale RGB-D dataset containing 4493 training scans and 549 validation scans.
In our experiments, the training scans are leveraged as the unlabeled input.
The validation scans are used for testing, whose ground-truth layouts are annotated by crowd-sourcing.
% We employ the training split as an unlabeled data source to test the utility of our methodology on the validation set with crowdsourced layout quad annotation.

\textbf{Implementation Details}\quad
In the transformation stage, the point cloud is first downsampled to 40,000 points with FPS and rotated along the z-axis by $\theta = \theta_1 + \theta_2$, with $\theta_1$ randomly chosen from $\{0, \frac{\pi}{2}, \pi, \frac{3\pi}{2}\}$ and $\theta_2$ uniformly sampled from $[-5^{\circ}, 5^{\circ}]$. 
Next, the point cloud is flipped along the x-axis and the y-axis with the probability of 0.5 and scaled by a ratio uniformly sampled from $[0.85, 1.15]$. 

% For data transformation, we set $K = 40,000$, $\theta = \theta_1 + \theta_2$, where $\theta_1$ is randomly chosen from $\{0, \frac{\pi}{2}, \pi, \frac{3\pi}{2}\}$ and $\theta_2$ subjects to the uniform distribution $U[-5^{\circ}, 5^{\circ}]$. $R$ is also uniformly sampled from $[0.85, 1.15]$.
% As for mean teacher ramp-up epochs, we follow \cite{zhao2020sess}.
We implement the teacher and student models in the proposed Quad Mean Teacher framework with PQ-Transformer \cite{chen2022pq}, while the framework also works with other layout estimators.
The preprocessing of quad annotations and the evaluation metrics are the same as \cite{chen2022pq}.
The consistency loss weight is set to $\lambda_\text{QMT} = 0.05$, using the same warm-up strategy as \cite{zhao2020sess}.
% Following \cite{zhao2020sess}, the consistent loss weight is set as $\lambda_\text{CMT} = 0.05 \times \exp(-5 * (1-(i / 1000)^2))$ at $i$-th iteration. \red{????}
The pseudo-label loss weight is set as $\lambda_\text{GMF}=5\times10^{-4}$. 
Our experiments run on a single NVIDIA RTX A4000 GPU with batch size of 6.
Half of the samples in a batch have quad annotations.
% The experiments run with a single NVIDIA RTX A4000 GPU.

\subsection{Results}
% As far as we know, there are no prior works sharing the same semi-supervised setting of layout estimation task with us.
% Hence we make comparisons to fully-supervised models on layout estimation task.
To the best of our knowledge, our methods are the first to perform the PC-based layout estimation task in the semi-supervised and the omni-supervised setting.
Hence we compare our method with fully-supervised methods including SceneCAD \cite{avetisyan2020scenecad} and PQ-Transformer \cite{chen2022pq}.
% SceneCAD \cite{avetisyan2020scenecad} adopts a bottom-up approach to predict the room layout, which predicts corners, edges, and planar elements sequentially.
% And PQ-Transformer \cite{chen2022pq} straightforwardly proposes quad proposals and refines them with the ability of Transformer.
% We add our consistency loss and pseudo-label refinement module based on the code of former state-of-the-art model \cite{chen2022pq}.
% The specification of all preprocessing techniques of quad annotations and thresholds of evaluation metrics remains the same with \cite{chen2022pq}.

% Unless specified, all other implementation details follow PQ-Transformer \cite{chen2022pq}.

\begin{table}[htbp]
\centering
\caption{Layout Estimation F1-scores on ScanNet}
\begin{center}
\resizebox{0.9\columnwidth}{!}{

\newcolumntype{Y}{>{\centering\arraybackslash}X}
\begin{tabularx}{1.00\linewidth}{cYYYYYY}
\toprule
Method & 5$\%$ & 10$\%$ & 20$\%$& 30$\%$& 40$\%$& 100$\%$ \\
\midrule
SceneCAD \cite{avetisyan2020scenecad} & - & - & - & - & - & 37.90\\
PQ-Transformer \cite{chen2022pq} & 22.43 & 29.26 & 39.60 & 46.02 & 48.08 & 56.64 \\
\midrule
Ours (QMT only) & 26.83 & 34.76 & 44.42 & 49.30 & 51.84 & 58.80\\

Ours (GMF only) & 26.65 & 35.17 & 45.25 & 51.69 & 52.69 & 60.54 \\

\textbf{Ours (QMT \& GMF)} & \textbf{29.08} & \textbf{36.85} & \textbf{48.68} & \textbf{54.35} & \textbf{56.92} & \textbf{60.75} \\
\midrule
Margin & +6.65 & +7.59 & +9.08 & +8.33 & +8.84 & +4.11 \\
Relative Improv. ($\%$) &  29.65$\uparrow$ & 25.94$\uparrow$ & 22.93$\uparrow$ & 18.10$\uparrow$ & 18.37$\uparrow$ & 7.26$\uparrow$ \\
\bottomrule
\end{tabularx}
}
\label{tab:layout}
\end{center}
\vspace{-10pt}
\end{table}

% \textbf{Results.} We consider two task settings, one semi-supervised on ScanNet, and the other omni-supervised towards ARKitScenes.

We evaluate our method and the baselines in various semi-supervised settings on ScanNet validation set and report the F1-scores of the predicted layouts in Tab.~\ref{tab:layout}.
The size of labeled set $\mathcal{D}_L$ sampled from the ScanNet training split, or the amount of ground truth annotations in use, is denoted by percentages in the first row. 
And $\mathcal{D}_U$ is the complementary set whose annotations are assumed unknown.

% 这个在 intro 里提过...
It can be seen that either QMT or GMF can result in performance boost. And by combining these two techniques together, we see further improvement in performance.
In all semi-supervised settings, the performances of our methods are better than baselines by large margins.
With only 40$\%$ quad annotations available, our method achieves similar performance to that of the state-of-the-art method trained in fully supervised settings.
Surprisingly, our method also performs better in fully supervised settings than former arts.
We attribute the outperformance to the consistency regularization mechanism promoting the model's robustness to perturbations and the pseudo-label refinement module providing guidance on the geometrical information of layouts.

% A varying ratio of ScanNet training split is considered as training input for former arts. We report the performance of trained model on the whole ScanNet validation set. As for our method, we leverage the same data as the labeled dataset in conjunction with the whole ScanNet training split as a unlabeled one (See Fig. \ref{fig:main}). We present results with only our Confident Mean Teacher (CMT) module, only our Gamma Mixture Filtering (GMF) module and a mixture of both of them (Ours).
% It can be seen that either CMT or GMF can function efficiently and significantly improve performance by a large margin. Furthermore, by combining these two proposed approaches, massive improvements continue to exist under any possible ratio setting. With only 40$\%$ of labeled data, we could get comparable outcomes with a former fully-supervised one. And in a fully-supervised setting, we continue to boost former SOTA by a margin of +4.11$\%$.

We further demonstrate the robustness of our method in the omni-supervised setting \cite{radosavovic2018data}.
To be more specific, we train our method and the baselines with the whole labeled training split of ScanNet $\mathcal{D}_L$ and then evaluate the performance on the validation split of the ARKitScenes dataset with crowd-sourced layout annotations.
Besides, in our method, the unlabeled training split of ARKitScenes serves as the unlabeled dataset $\mathcal{D}_U$.
As shown in Tab.~\ref{tab:arkit}, our method achieves a significant margin over former arts, showing the ability to generalize to more realistic omni-supervised settings.

\begin{figure}[t]
  \centering
  \includegraphics[width=1.00\linewidth]{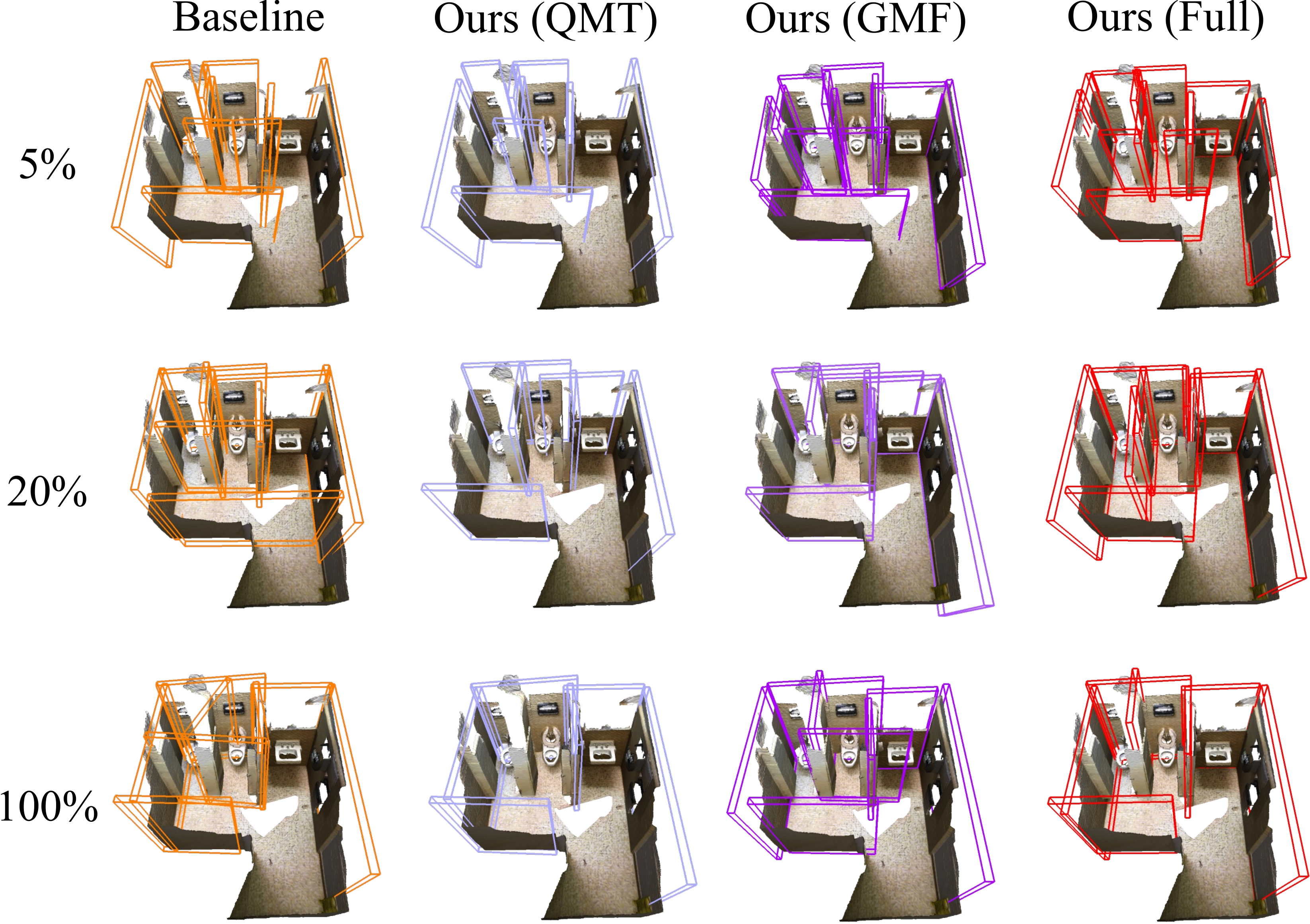}
  \caption{\textbf{Qualitative results on ScanNet.} The ratio represents the proportion of annotated data in use.}
  \label{fig:comparison} 
  \vspace{-10pt}
\end{figure}

\begin{figure}[t]
  \centering
  \includegraphics[width=0.9\linewidth]{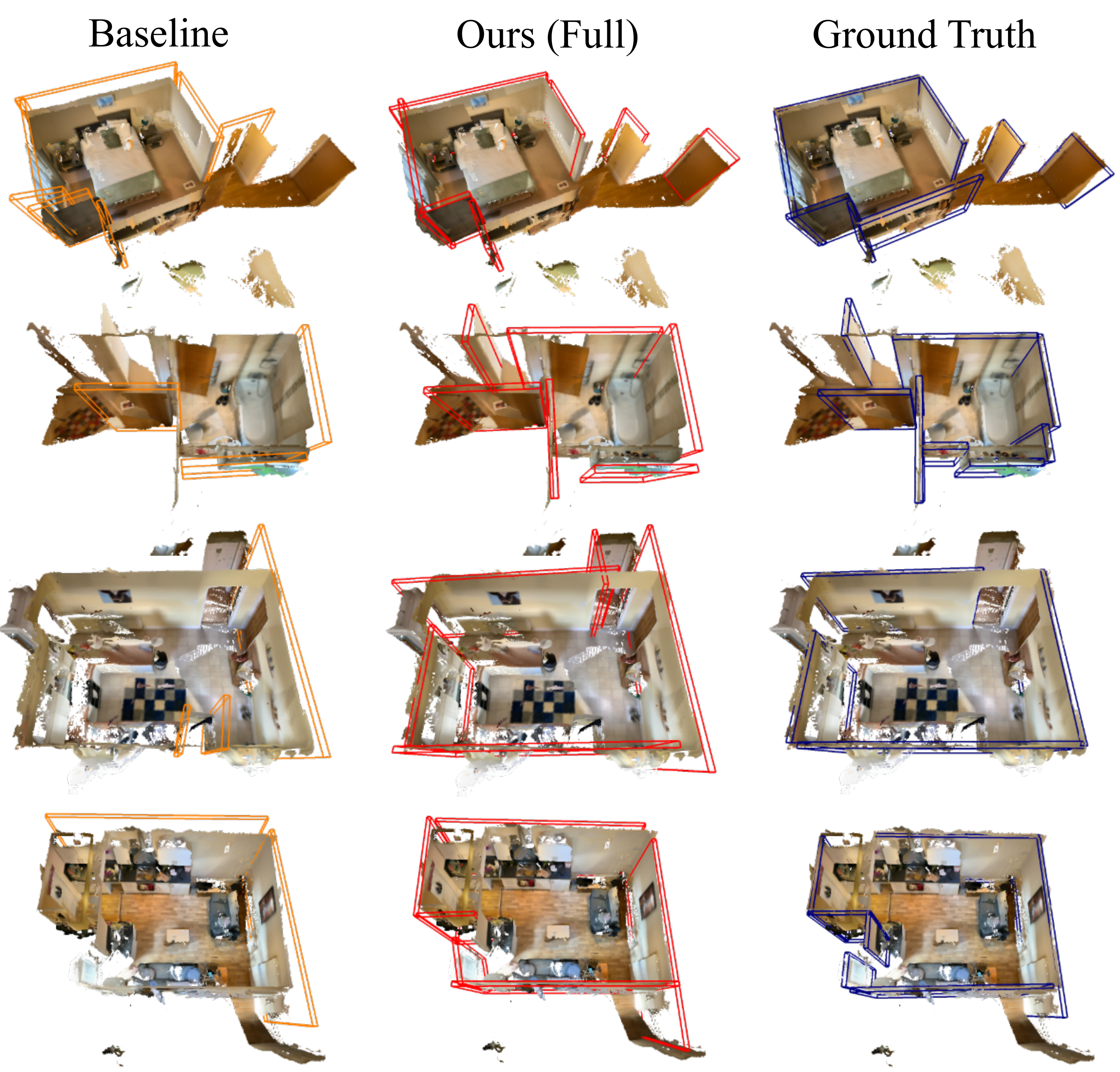}
  \caption{\textbf{Qualitative results on ARKitScenes.} Ground truth layouts are annotated by crowd-sourcing.}
  \label{fig:arkit}
\end{figure}

In addition, we provide visualization of the quad predictions of our method on ScanNet and ARKitScenes in Fig. \ref{fig:comparison} and Fig. \ref{fig:arkit}.
These qualitative results show that exterior quads as well as the interior quads are predicted by our method accurately, compensating for the ineffectiveness of PQ-Transformer \cite{chen2022pq} w.r.t. interior wall quads.

\begin{table}[htbp]
\caption{Layout Estimation F1-scores on ARKitScenes}
\begin{center}
\resizebox{0.9\columnwidth}{!}{
\begin{tabular}{c|cc|c}
\toprule
Method & Recall (\%) & Precision (\%) & F1-score (\%) \\
\midrule
%3D-SIS & \multicolumn{1}{|c|}{Geo + 5 views} & 40.2 & 22.5  \\
PQ-Transformer \cite{chen2022pq} & 6.72 & 25.81 &  10.66  \\
%Group-Free(L12, O512) & \multicolumn{1}{|c|}{Geo only} & 69.1(68.6) & 52.8(51.8)  \\
\midrule 
\textbf{Ours (QMT \& GMF)} & \textbf{23.00} & \textbf{29.50} & \textbf{25.85} (+15.19) \\
\bottomrule
\end{tabular}
}
\label{tab:arkit}
\end{center}
\vspace{-10pt}
\end{table}

\subsection{Ablation Study}

\textbf{Data Transformation Strategies}
% To investigate the necessity of data transformation,
We run the 10$\%$-supervised experiment on ScanNet with different data transformations.
As shown in Tab.~\ref{tab:augmentation}, data transformation is crucial to our proposed method, as any of the transformations improves the performance, and in extreme cases without transformations the F1-score decreases by 6.31\%.
% In the extreme case that we remove all transformation schemes, the performance drops dramatically by 6.31$\%$. 

Among the four transformations, rotation has the largest influence on the performances.
One possible reason is that rotation brings the most changes to the coordinates of points whilst keeping the holistic layouts of the scenes unchanged.
% Of all the four perturbation techniques, rotation plays a more vital role, as its activation is associated with the greatest increase from baseline and its deactivation with the greatest reduction compared to the original framework.

\begin{table}[htbp]
\caption{Ablations on Data Transformation Strategies}
\begin{center}
\resizebox{0.9\columnwidth}{!}{
\begin{tabular}{cccc|c}
\toprule
 Downsample & Flipping & Rotation & Scaling & F1-score (\%)\\
\midrule
    $\times$  & $\times$ & $\times$ & $\times$ & 30.54\\
\midrule
    \checkmark & $\times$ & $\times$ & $\times$ & 31.59\\
   $\times$ & \checkmark & $\times$ & $\times$ & 32.47\\
   $\times$ & $\times$  & \checkmark & $\times$ & 32.95\\
   $\times$ & $\times$  & $\times$  & \checkmark & 30.70\\
\midrule
   $\times$ & \checkmark & \checkmark & \checkmark & 35.61\\
  \checkmark & $\times$ & \checkmark & \checkmark & 34.53\\
  \checkmark & \checkmark & $\times$ & \checkmark & 33.42\\
  \checkmark & \checkmark & \checkmark & $\times$ & 33.96\\
\midrule
  \checkmark & \checkmark & \checkmark & \checkmark & \textbf{36.85}\\

\bottomrule 
\end{tabular}
}
\label{tab:augmentation}
\end{center}
\end{table}

\textbf{Quad Mean Teacher}\quad
We compare Quad Mean Teacher and the basic Mean Teacher (MT) method in the 10\%-supervised settings and report their performances on ScanNet in Tab.~\ref{tab:meanteacher}.
MT assumes that all teacher predictions are equally reliable.
% We look into the impact of confidence factor in consistency loss.
% We repeat the 10$\%$-supervised experiment on ScanNet with consistency loss regularization alone, and the Confident Mean Teacher (CMT) is swapped out for a standard Mean Teacher module, which indeed assumes the premise that all confident scores are $1.0$, namely all predictions of teachers are expected to be trusted.
Results show that the QMT achieves a large margin over MT on the precision of prediction.
% Result shows that substituting the CMT for a standard one leads to a noticeable performance drop.
We believe this is because the confidence of predictions by the teacher model is exploited and erratic or incorrect predictions are neglected accordingly.

% We believe this is because in our method, erratic or incorrect predictions are neglected to the extent of the teacher's self-awareness of the reliability .

\begin{table}[htbp]
\caption{Ablations on Quad Mean Teacher}
\begin{center}
\resizebox{0.9\columnwidth}{!}{

\begin{tabular}{c|cc|c}
\toprule
Method & Recall (\%) & Precision (\%) & F1-score (\%) \\
\midrule
Mean Teacher & 26.39 & 39.61 & 31.67\\
\midrule 
\textbf{Ours (QMT)} & \textbf{27.73} & \textbf{46.54} & \textbf{34.76}  \\
\bottomrule
\end{tabular}
}

\label{tab:meanteacher}
\end{center}
\vspace{-10pt}
\end{table}

\textbf{Gamma Mixture Filtering}
% We examine the effects of Gamma Mixture Filtering (GMF) that is previously presented in Section \ref{GMF}.
In the 10\%-supervised settings, we compare our method using only pseudo-label loss with the naive $
\epsilon_D$ approach introduced in Sec.~\ref{GMF}. We set the fixed threshold $\epsilon_D = 0.2 \text{m}$.
% We run the 10$\%$-supervised experiment on ScanNet with pseudo-label loss alone, with GMF strategy replaced with a simply $\epsilon_D = 0.2 \text{m}$ filtering introduced in the same section. 
More specifically, in the alternative method, a point stays after filtering if its perpendicular distance to the plane is less than $\epsilon_D$.
Compared to ours, the alternative method achieves significantly lower performance, since no supervision is applied on the quad normals and sizes.

% No appreciable performance increase is seen in the latter method, as it comes little aid to the refinement of quad normals and sizes.

\begin{table}[htbp]
\caption{Ablations on Gamma Mixture Filtering}
\begin{center}
\begin{tabular}{c|cc|c}
\toprule
Method & Recall (\%) & Precision (\%) & F1-score (\%) \\
\midrule
Simple $\epsilon_D$ Filtering & 25.29 & 40.51 & 31.14\\
\midrule
\textbf{Ours (GMF)} & \textbf{29.75} & \textbf{43.01} & \textbf{35.17}  \\
\bottomrule
\end{tabular}
\label{tab:gmf}
\end{center}
\vspace{-10pt}
\end{table}

\section{Conclusion}
% In this research, we introduce a method for semi-supervised estimation of room layout quads using point clouds.
Our research makes the first step towards omni-supervised layout estimation merely using point clouds, which has promising implications in robotics.
Our training framework combines Quad Mean Teacher and Gamma Mixture Filtering to better exploit the unlabeled data.
% The unique combination of Confident Mean Teacher and Gamma Mixture Filtering, which respectively regularizes consistency loss and harvests robust pseudo-labels, is what makes our work successful.
Experimental results demonstrate our method's effectiveness in semi-supervised, fully-supervised and omni-supervised settings.
% With 40$\%$ of layout quad annotations available, our method outperforms fully-supervised prior arts.
% In fully-supervised and omni-supervised settings, our method continues to achieve margins over strong baselines.

Despite the effectiveness of our method, limitations still exist.
The predictions of our method are unsatisfactory in incomplete scenes, in which insufficient points fail to form a layout wall.
In the future, we will consider possible rectifications including ensembling online inference results, thanks to the quasi-real-time speed brought by the PQ-Transformer \cite{chen2022pq} implementation.
% Our research takes a significant first step toward reliable semi-supervised layout estimation merely using point clouds, which has fascinating implications in robotic sensing and motion planning.


%%%%%%%%%%%%%%%%%%%%%%%%%%%%%%%%%%%%%%%%%%%%%%%%%%%%%%%%%%%%%%%%%%%%%%%%%%%%%%%%

\bibliographystyle{IEEEtran}
\bibliography{References}

\end{document}